\documentclass[preprint,12pt]{elsarticle}

\usepackage{amsmath,lineno, url} 
\usepackage{multirow}
\usepackage{doi}
\usepackage{nomencl}
\usepackage{soul}
\usepackage{amssymb}
\usepackage{algorithm}
\usepackage{algorithmic}
\usepackage{graphicx}
\usepackage{hyperref} 
\usepackage{xcolor}
\usepackage{colortbl}
\soulregister{\citep}7
\makenomenclature

\journal{Smart Agricultural Technology}

\begin{document}
\begin{frontmatter}

\title{Tree-SLAM: semantic object SLAM for efficient mapping of individual trees in orchards}

\author[inst1]{David Rapado-Rincon}

\affiliation[inst1]{organization={Agricultural Biosystems Engineering, Wageningen University \& Research},
            addressline={P.O. Box 16}, 
            city={Wageningen},
            postcode={6700 AA}, 
            country={The Netherlands}}
\author[inst1]{Gert Kootstra}

\begin{abstract}
Accurate mapping of individual trees is an important component for precision agriculture in orchards, as it allows autonomous robots to perform tasks like targeted operations or individual tree monitoring. However, creating these maps is challenging because GPS signals are often unreliable under dense tree canopies. Furthermore, standard Simultaneous Localization and Mapping (SLAM) approaches struggle in orchards because the repetitive appearance of trees can confuse the system, leading to mapping errors. To address this, we introduce Tree-SLAM, a semantic SLAM\nomenclature{SLAM}{Simultaneous Localization and Mapping} approach tailored for creating maps of individual trees in orchards. Utilizing RGB-D\nomenclature{RGB-D}{Red, Green, Blue - Depth} images, our method detects tree trunks with an instance segmentation model, estimates their location and re-identifies them using a cascade-graph-based data association algorithm. These re-identified trunks serve as landmarks in a factor graph framework that integrates noisy GPS\nomenclature{GPS}{Global Positioning System} signals, odometry, and trunk observations. The system produces maps of individual trees with a geo-localization error as low as 18 cm, which is less than 20\% of the planting distance. The proposed method was validated on diverse datasets from apple and pear orchards across different seasons, demonstrating high mapping accuracy and robustness in scenarios with unreliable GPS signals.
\end{abstract}

\begin{keyword}
semantic SLAM, agricultural robotics, multi-object tracking, factor graph
\end{keyword}

\end{frontmatter}
\printnomenclature

\section{Introduction}
A significant decline in available agricultural labor presents a challenge for sustaining agricultural production, potentially leading to food losses \cite{kootstra_robotics_2020, kootstra_selective_2021}. This challenge, coupled with the growing need for more sustainable farming practices, highlights the importance of technological solutions. Automation and robotics are emerging as key technologies to address these issues, offering the potential to enhance productivity, by compensating for labor scarcity and optimizing farm management through data-driven insights \cite{sparrow_robots_2021, kootstra_advances_2024}. This is particularly relevant in high-value crops such as those found in orchards, where precise operations have the potential to improve efficiency and reduce labor needs.

For autonomous robots to perform tasks effectively in orchards, such as targeted spraying or individual tree monitoring, they require a detailed map of the environment and the ability to determine their position within it. This dual capability of mapping and localization is fundamental for enabling precise operations at the tree level, which is essential for optimizing yield and resource usage \cite{dong_semantic_2020, tiozzo_fasiolo_towards_2023}. While Real-Time Kinematic Global Positioning System (RTK-GPS)\nomenclature{RTK-GPS}{Real-Time Kinematic Global Positioning System} sensors are commonly used for positioning due to their high accuracy under open sky conditions, their performance significantly degrades in environments with dense canopy cover, such as orchards \cite{shalal_review_2013, tiozzo_fasiolo_towards_2023}. This limitation necessitates the use of alternative methods like sensor fusion or Simultaneous Localization and Mapping (SLAM), which can provide reliable positioning and mapping in such challenging conditions by building a map while simultaneously tracking the robot's location within it \cite{tiozzo_fasiolo_towards_2023}.

Traditional vision-based SLAM algorithms using low-level image features, however, struggle in homogeneous environments like orchards and vineyards, where visual features are repetitive or lack distinctiveness \cite{hroob_benchmark_2021, tiozzo_fasiolo_towards_2023}. In agricultural settings, many plants or trees appear visually similar, making it difficult for feature-based methods to distinguish between different locations and reliably track the robot's pose. Moreover, the uniformity of crop rows can lead to perceptual aliasing, causing the SLAM system to incorrectly associate observations with previously mapped areas, leading to accumulated errors over time. Recent developments in semantic SLAM, which use object detection to build maps based on recognized objects rather than low-level features, have shown better performance than traditional vision SLAM methods in GPS-restricted outdoor environments \cite{shao_monocular_2021}. Semantic SLAM methods leverage recognizable objects as landmarks, which requires a robust data association mechanism to match new detections to previously observed landmarks. This process of re-identification is fundamental for maintaining a consistent map. Many of these systems use a factor graph to fuse sensor measurements and optimize the estimated positions of the robot and the landmarks. 

Since orchards contain static, identifiable objects like tree trunks, a semantic object SLAM approach appears well-suited for these environments. However, many existing semantic SLAM approaches are designed for complex 3D environments \cite{nicholson_quadricslam_2019, shao_monocular_2021}, which introduces computational overhead that might be unnecessary for many agricultural applications using unmanned ground vehicles (UGVs). Orchard mapping, for instance, can often be simplified to a 2D representation, offering significant gains in efficiency. Furthermore, these methods often do not explicitly model the relationships between static objects, such as the fixed positions of trees in a row, which can be a valuable source of information for robust localization and mapping.

In this paper, we present Tree-SLAM, a semantic SLAM approach specifically designed for orchard environments, aiming to map and geo-localize individual trees. Our method detects tree trunks in RGB-D images using an instance segmentation model and employs a cascade graph-based data association algorithm to re-identify these trunks across different viewpoints. These re-identified trunks serve as landmarks in our factor graph based SLAM framework, which integrates multiple sensor measurements, including odometry, GPS (when available), and observations of tree trunks. We evaluated the robustness of this solution in real-world orchard scenarios.

Our contributions are:
\begin{itemize}
    \item A large tree trunk instance segmentation dataset, containing more than 8,000 images from apple and pear trees with different ages across different dates, seasons, and variable weather conditions.
    \item A cascade graph based data association algorithm that uses graph-based neighborhood relationships to accurately associate tree trunks over time.
    \item A multi-sensor semantic object SLAM method based on a factor graph that integrates GPS and odometry localization measurements with semantic RGB-D camera-based object detections.
    \item An extensive evaluation of the proposed system in real-world orchard conditions, including ablation studies to validate the contribution of individual components.
\end{itemize}

\section{Materials and methods}
\subsection{Robotic setup}
\begin{figure}[ht]
    \centering
    \includegraphics[width=0.48\textwidth]{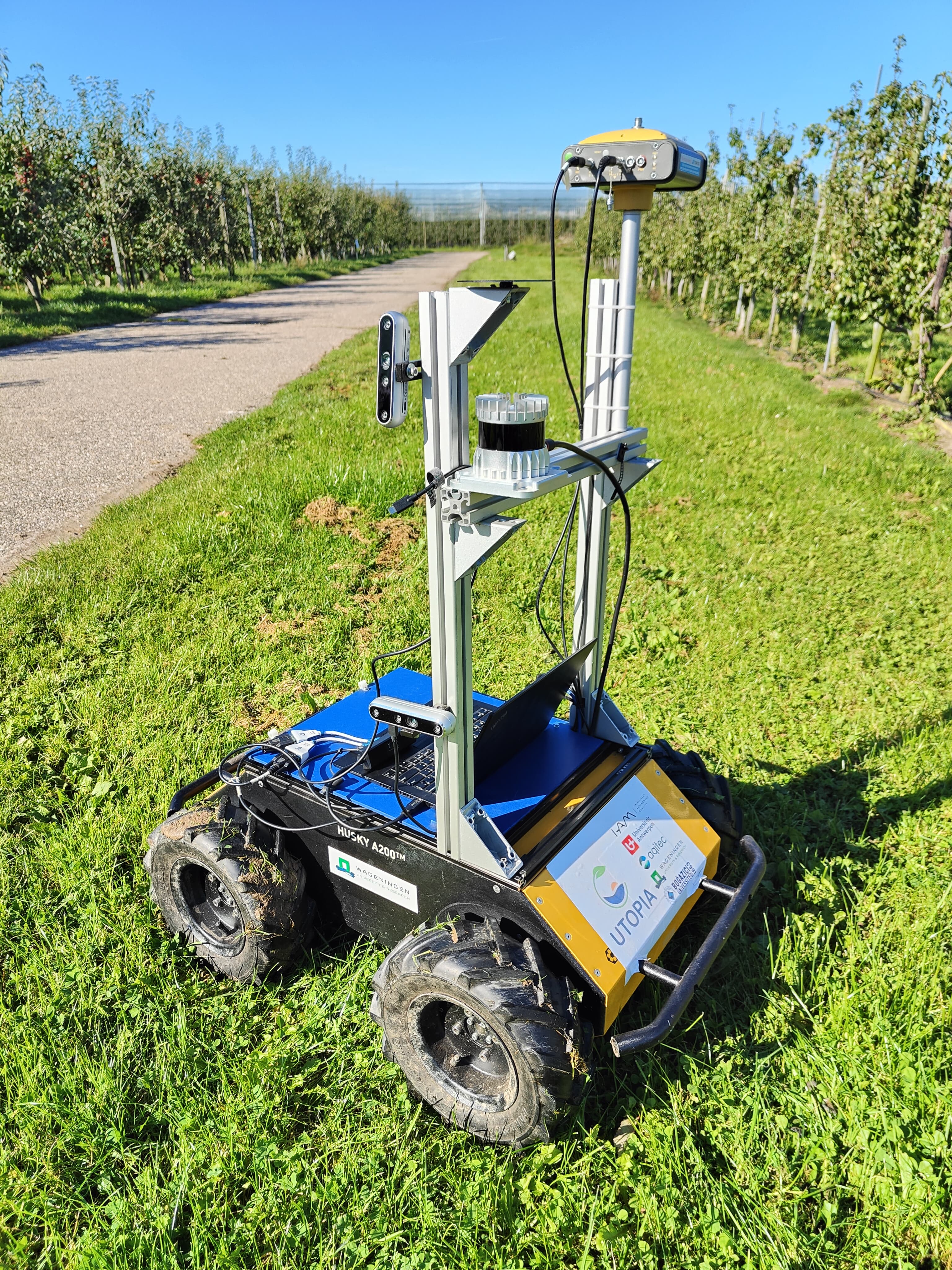}
    \caption{Robotic system used for data collection. The robot carries two RealSense D455 RGB-D cameras, from which only the bottom camera was used in this work. The robot is equipped with a Topcon HiPer Pro RTK-GPS receiver, an Xsens MTi-300 IMU and an Ouster OS-0 LiDAR. The IMU device is occluded behind the LiDAR. The LiDAR was not used in this work. The robot is shown in the orchard environment where the data was collected.}
    \label{fig:robot}
\end{figure}

We developed a UGV \nomenclature{UGV}{Unmanned Ground Vehicle} system using the Clearpath Husky platform, as shown in Figure \ref{fig:robot}. The robot was equipped with a Topcom HiPer Pro RTK-GPS receiver operating at 10 Hz, an Xsens MTi-300 \nomenclature{IMU}{Inertial Measurement Unit} operating at 400 Hz, and two Intel RealSense D455 RGB-D cameras at a resolution of 1280x720 and 15 Hz. The IMU is only used to detect turns at the end of the row. Both cameras were mounted facing the right side of the robot. Only the bottom camera was used for this work as it was placed at a similar height as the tree trunks. Robot Operating System 2 (ROS2\nomenclature{ROS2}{Robot Operating System 2}) \cite{macenski_robot_2022} was used to control the UGV and the sensors. To create a transformation graph of the robot, camera to IMU calibration was performed using Kalibr \cite{rehder_extending_2016}, while GPS to IMU transformation was measured.

\subsection{Data acquisition}
Two datasets were collected during this work, one dataset to train an instance segmentation algorithm, and a second dataset to perform and evaluate mapping in apple and pear orchards. The instance segmentation dataset was collected over several days across October 2023, December 2023, February 2024, and September 2024. It resulted in a total of 7284 train and 395 validation images. The tree trunks in the images were labeled for instance segmentation.

\begin{table}[htbp]
\centering
\caption{Overview of the mapping dataset collected from four different orchard rows. For each row, the table details the tree type, planting distance, number of ground truth (GT) trees measured, data collection dates, test set size for instance segmentation, and the robot's trajectory type (Path). For the dates, we group them into Leafless and Leafed based on the presence or absence of foliage.}
\resizebox{\linewidth}{!}{
    \begin{tabular}{ccccccc} 
    \hline
    \textbf{Row} & \textbf{Type} & \textbf{Planting Distance} & \textbf{\# GT trees measured} & \textbf{Date} & \textbf{Detection test set images} & \textbf{Path}\\ \hline
    \multirow{2}{*}{Pear 1} & \multirow{2}{*}{Pear} & \multirow{2}{*}{1.1 m} & \multirow{2}{*}{32 of 135} & 28/02/2024 (Leafless) & 126 & U-shape\\ 
    &  &  &  & 31/10/2023 (Leafed) & 160 & U-shape\\ \hline
    \multirow{2}{*}{Pear 2} & \multirow{2}{*}{Pear} & \multirow{2}{*}{1.1 m} & \multirow{2}{*}{32 of 135} & 28/02/2024 (Leafless) & 127 & U-shape\\ 
    &  &  &  & 20/9/2024 (Leafed) & 124 & U-shape\\ \hline
    \multirow{2}{*}{Pear 3} & \multirow{2}{*}{Pear} & \multirow{2}{*}{1.1 m} & \multirow{2}{*}{32 of 135} & 28/02/2024 (Leafless) & 192 & U-shape\\ 
    &  &  &  & 20/9/2024 (Leafed) & 154 & U-shape\\ \hline
    \multirow{2}{*}{Apple 1} & \multirow{2}{*}{Apple} & \multirow{2}{*}{1.2 m} & \multirow{2}{*}{65 of 65} & 28/02/2024 (Leafless) & 200 & Full\\ 
    &  &  &  & 31/10/2023 (Leafed) & 155 & Full\\ \hline
    \end{tabular}
}
\label{tab:dataset}
\end{table}

\begin{figure}[ht]
    \centering
    \includegraphics[width=\textwidth]{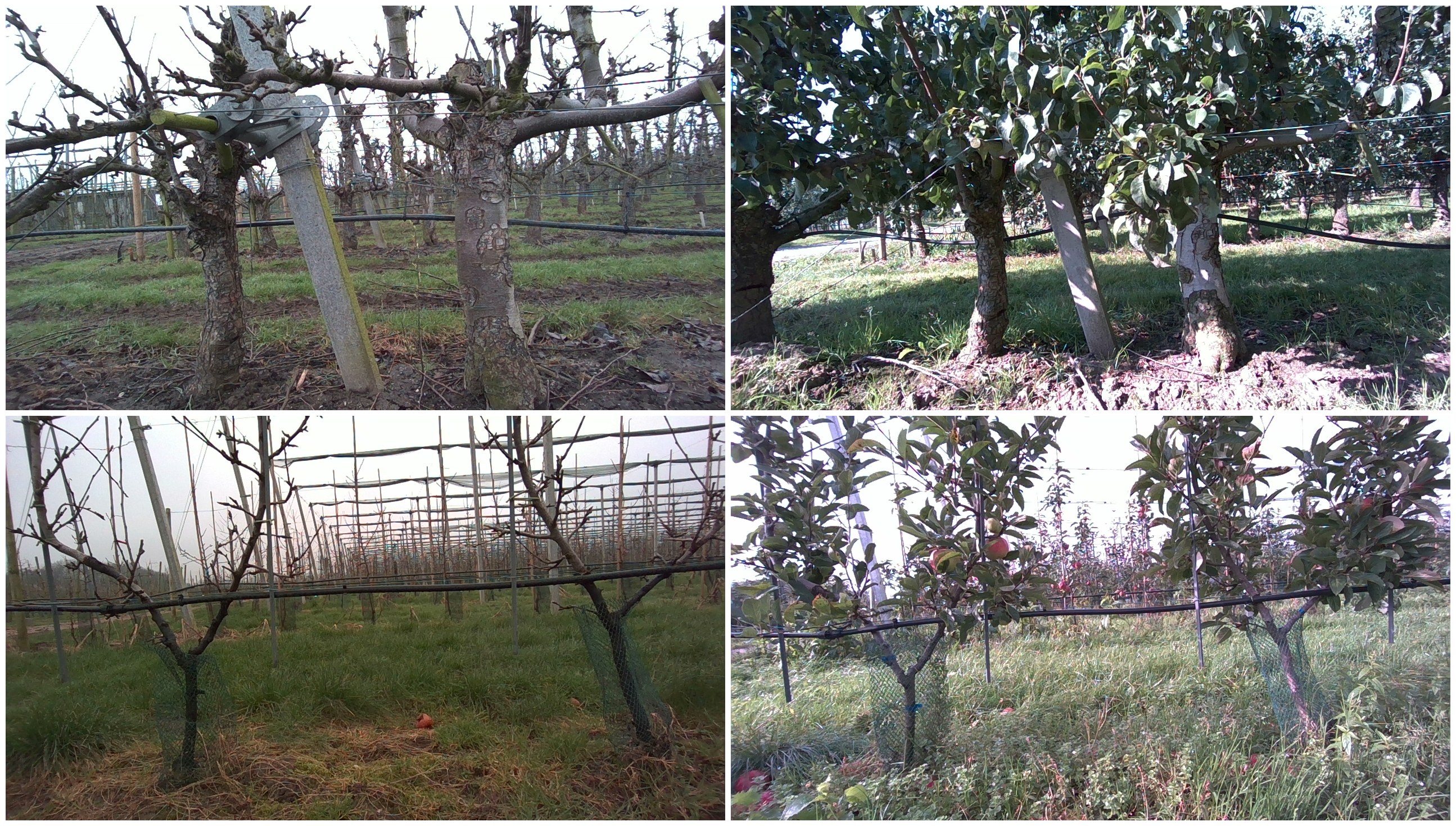}
    \caption{Examples of the pear (top) and apple (bottom) trees of the mapping dataset across leafless (left) and leafed (right) conditions. Leafless and leafed images contain the same trees. The apple trees are younger than the pear trees. Furthermore, they have a protective net around the trunk. Tall grass might cause occlusions in the apple trees in leafed conditions.}
    \label{fig:tree_examples}
\end{figure}

Similarly, the mapping dataset was collected across different dates of the year as shown in Table \ref{tab:dataset}. We categorize the recordings as "Leafless" or "Leafed" according to the presence or absence of foliage. To collect data from a row, the robot was driven around both sides of the row of trees, either forming a U-shaped pattern or a full loop around the row. The U-shaped recordings are a result of hardware limitations of the laptop controlling the robot, where multiple sensor messages were missed if a full loop was performed in the pear orchards with 135 trees. In the U-shaped paths, the robot started at some point in the middle of the row, drove toward an end, made a 180-degree turn and drove on the other side of the row until the first tree seen at the beginning was seen again from its opposite side. In the full loop path, the robot started and ended in roughly the same position on the row, after driving through both sides of the row of trees and performing a 180-degree turn at both ends of the row. Some examples of trees from the collected rows can be seen in Figure \ref{fig:tree_examples}. The pear trees were older with thicker trunks while the apple trees were younger, thinner and with a rodent-protection net around the trunk.

For each row, the ground truth GPS location of multiple trees was recorded using a GPS stick using RTK corrections. The stick was placed at two opposite sides of the tree trunk, and the coordinates were averaged between both measurements. Table \ref{tab:dataset} describes the different rows collected and the number of trees measured per row.

\subsection{Mapping algorithm}
\begin{figure}[ht]
    \centering
    \includegraphics[width=\textwidth]{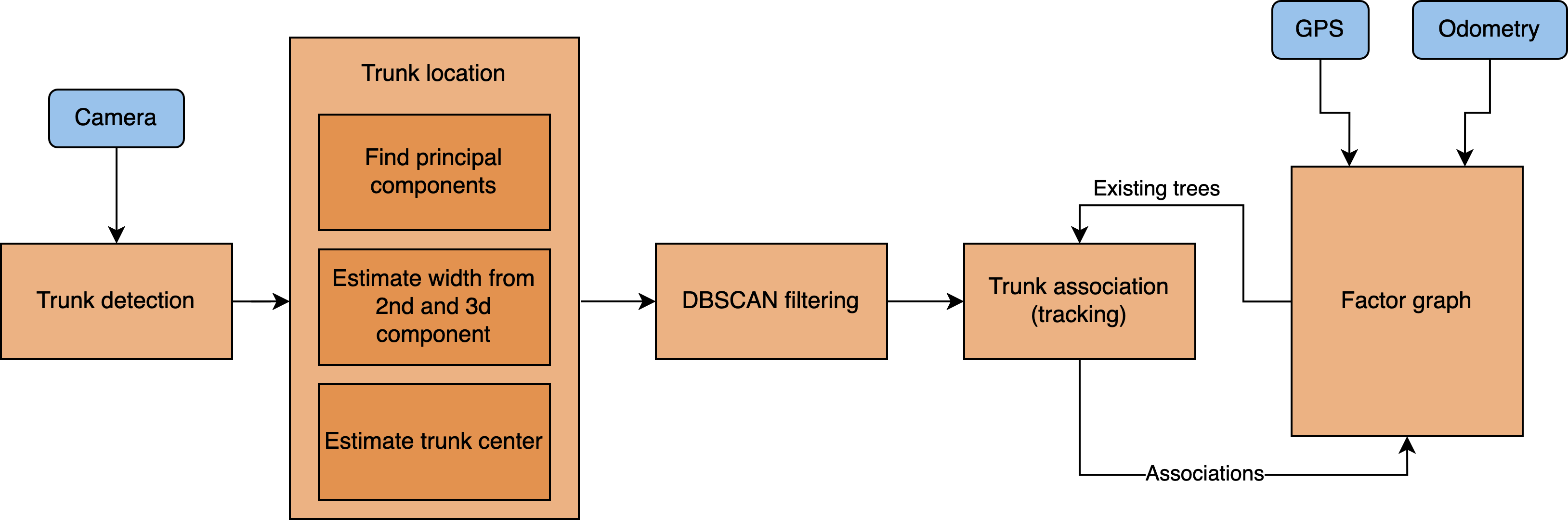}
    \caption{Overview of the Tree-SLAM pipeline, illustrating the main processing stages from sensor input to factor graph update. The factor graph feeds the tracking and association step with the known trees. Then, the associated detections are used to update the factor graph with the new observations. The process shown in this figure is repeated at each time step \( t \).}
    \label{fig:pipeline}
\end{figure}

This work proposes Tree-SLAM, a semantic SLAM approach tailored for orchard environments. As shown in Figure \ref{fig:pipeline}, the algorithm processes data at each time step \( t \). It begins with the object detection stage, where an instance segmentation model, YOLOv8 \cite{jocher_ultralytics_2023}, identifies tree trunks in color images. The resulting masks isolate the corresponding points in the point cloud generated by the RGB-D camera. This results in a point cloud for each detected trunk, enabling the estimation of its location relative to the robot. Subsequently, these detections are associated with previously identified trees (landmarks). Finally, the associated detections, along with the robot's pose estimate derived from GPS, odometry, or both, are incorporated into a factor graph that combines measurements to perform an uncertainty-driven estimation of both the robot trajectory and the tree locations. The following sections provide a detailed description of each stage in the Tree-SLAM pipeline.

\subsubsection{Trunk detection and location estimation}
Tree trunk detection and localization is the starting point of our landmark-based SLAM. We employed the YOLOv8-x model \cite{jocher_ultralytics_2023} for instance segmentation, chosen for its favorable balance between inference speed and accuracy. The model was trained on the custom dataset described earlier, incorporating data augmentation techniques: HSV color shifts, rotations, translations, scaling, shearing, flipping, and mosaic augmentation to enhance robustness. At inference time, the trained YOLOv8-x model processes the camera's color image to detect and segment tree trunks. These segmented trunks serve as the primary landmarks for Tree-SLAM.  

To obtain the 3D location of each detected trunk, we process the corresponding RGB-D data. First, a point cloud is generated from the RGB-D image. This point cloud is then masked using the segmentation mask provided by YOLOv8, isolating the points belonging to the detected trunk. We then apply Principal Component Analysis (PCA)\nomenclature{PCA}{Principal Component Analysis} to this segmented point cloud. The first principal component typically aligns with the main axis (height) of the trunk. The second and third principal components describe the trunk's cross-section. We estimate the trunk width by calculating the maximum range of points along these second and third principal components.

This estimated width is needed for refining the trunk's location. Since the point cloud represents the visible surface of the trunk, its centroid is offset from the true center. We correct the centroid position \( \mathbf{p}^{\text{3D}} = [c_x, c_y, c_z]^T \) by adjusting it outwards along the direction perpendicular to the trunk axis (approximated by the vector from the camera to the centroid) by a distance approximating the trunk radius \( r \approx \text{width}/2 \). This adjustment yields a more accurate 3D estimate of the trunk's center \( \mathbf{p}^{\text{3D}} \). Finally, this 3D location is projected onto the 2D ground plane by discarding the Z (height) component, resulting in the 2D trunk position \( \mathbf{p}^{\text{2D}}  = [p_x, p_y]^T \) relative to the robot.

To maximize the number of potential tree detections at each time step, we adopted a low confidence threshold of 0.1 for the detections generated by YOLO during our experiments. While this increases the recall of true tree trunks, it also increases the number of false positive detections (e.g., poles, dense weeds). We employ a filtering step based on spatial clustering to mitigate the false positives introduced by the low detection threshold, particularly detections of non-trunk objects like poles. This filtering is performed by clustering the estimated 2D trunk positions \( \mathbf{p} \) using the Density-Based Spatial Clustering of Applications with Noise (DBSCAN)\nomenclature{DBSCAN}{Density-Based Spatial Clustering of Applications with Noise} algorithm. DBSCAN groups points that are closely packed together, marking as outliers those points that lie alone in low-density regions. We set the neighborhood radius parameter, \( \varepsilon \), to 60\% of the known tree planting distance for the specific orchard row. This value was determined empirically to effectively separate distinct tree clusters while filtering out detections that do not conform to the expected planting pattern. When DBSCAN clusters two or more detections together, we select the detection with the highest confidence and discard the rest.

As a result of this combined detection, localization, and filtering process, we obtain a set of tree trunk detections at each time step \( t \). This set is denoted as \( \mathcal{D}^t = \{ d^{1,t}, d^{2,t}, \dots, d^{n_t,t} \} \), where \( n_t \) is the number of detections. Each detection \( d^{i,t} \) comprises the bounding box in the image frame \( d^{i,t}_{\text{bbox}} \) and the estimated 2D world position \( d^{i,t}_p \). The 2D position \( d^{i,t}_p \) of each detection is calculated using the robot pose at frame $t$.

\subsubsection{Trunk association (tracking)}
Tree-SLAM relies on tracking to maintain consistent identities for tree trunks across multiple frames, enabling the system to recognize when the same tree is observed again from different viewpoints or at different times. Relevant subtasks of tracking are data association, which matches current sensor detections to previously established tracks, and object motion prediction, which predicts where the objects will appear in future frames. In our approach, we assume that the 2D world location of each tree track remains static over time, while the bounding box of each trunk in the image frame is predicted using a Kalman filter motion model.

At each time step \( t \), we maintain a set of active tracks, where each track \( \tau \) represents a unique tree landmark. The goal is to associate the set of new detections \( \mathcal{D}^t \) with the set of existing tracks from the previous step, \( \mathcal{T}^{t-1} \). Each track \( \tau^{j,t-1} \in \mathcal{T}^{t-1} \) stores the bounding box from its last observation, \( \tau^{j,t-1}_{\text{bbox}} \), its estimated 2D world position, \( \tau^{j,t-1}_p \), as maintained by the factor graph, and its velocity in the X axis, \( \tau^{j,t-1}_{x_{\text{vel}}} \).

\begin{figure}[ht]
    \centering
    \includegraphics[width=\textwidth]{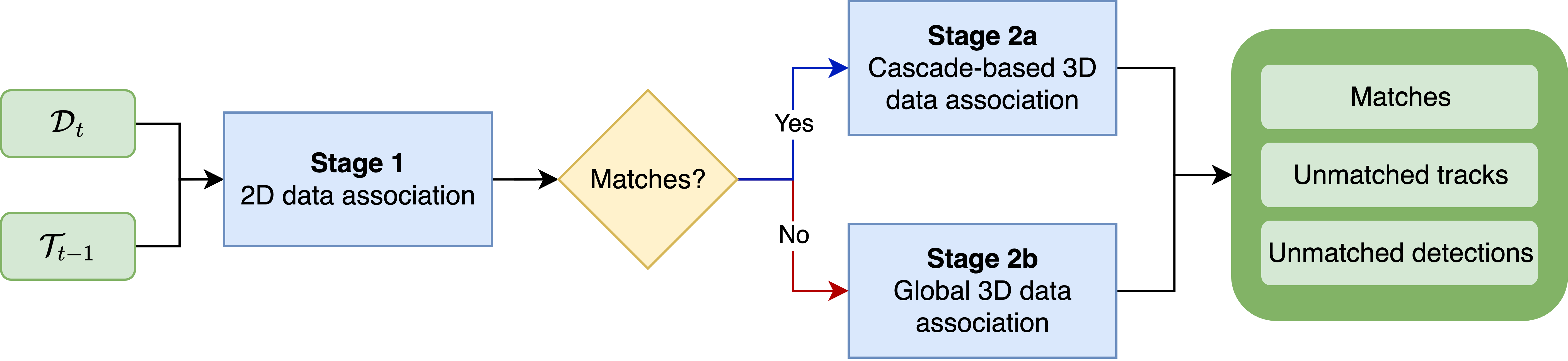}
    \caption{The trunk association and tracking algorithm. First a SORT-like association is performed in the image frame, matching predicted bounding boxes of existing tracks to new detections based on IoU. Unassociated detections and tracks are then processed in the world coordinate system using a cascade-graph approach that leverages spatial relationships between trees. If no associations are made in the first stage, a global association is performed using Euclidean distance.}
    \label{fig:association}
\end{figure}

The proposed data association algorithm, illustrated in Figure \ref{fig:association}, is a multi-stage process designed for robustness. It first attempts to match detections and tracks in the image space, then uses world coordinates and the spatial structure of the orchard to resolve ambiguities and match more difficult cases.

The first stage (Stage 1) performs association in the image frame. We adapt the approach from the Simple Online and Real-time Tracking (SORT)\nomenclature{SORT}{Simple Online and Real-time Tracking} method \cite{bewley_simple_2016}. For each active track \( \tau^{j,t-1} \) seen within the last five frames, the velocity of each track is used to predict its bounding box position in the current camera frame. We then compute a cost matrix between these predicted boxes and the new detection boxes \( d^{i,t} \in \mathcal{D}^t \) based on the Intersection-over-Union (IoU)\nomenclature{IoU}{Intersection-over-Union} distance. The Hungarian algorithm solves this assignment problem, providing an initial set of matched tracks and detections. The IoU cost is defined as:
\begin{equation}
C_{\text{IoU}}(i, j) = 1 - \text{IoU}( \tau^{j,t|t-1}_{\text{bbox}}, d^{i,t}_{\text{bbox}} )
\end{equation}
where \( \tau^{j,t|t-1}_{\text{bbox}} \) is the predicted bounding box of track \( \tau^{j,t-1} \), and \( d^{i,t}_{\text{bbox}} \) is the bounding box of detection \( d^{i,t} \).

For any detections \( \mathcal{D}^t_{\text{unassoc}} \) and tracks \( \mathcal{T}^{t-1}_{\text{unassoc}} \) that remain unassociated, we proceed to the second stage (Stage 2a), which operates in the 2D world coordinate system. This stage uses a cascade graph approach that leverages the known spatial arrangement of trees:

\begin{enumerate}
    \item \textbf{Initialization:} Start with the set of confidently matched tracks from the IoU association, \( \mathcal{T}^t_{\text{matched}} \). If this set is empty, we skip to a global association step described below.
    \item \textbf{Neighbor-based Association:} For each matched track \( \tau^{j,t} \in \mathcal{T}^t_{\text{matched}} \), we identify its neighboring unassociated tree tracks \( \mathcal{N}_{\tau^{j,t}} \) within a radius \( r \) in the 2D world coordinate system:
    \begin{equation}
    \mathcal{N}_{\tau^{j,t}} = \{ \tau^{k,t-1} \in \mathcal{T}^{t-1}_{\text{unassoc}} \mid \| \tau^{j,t}_p - \tau^{k,t-1}_p \| < r \}.
    \end{equation}
    We then attempt to match the tracks in \( \mathcal{N}_{\tau^{j,t}} \) with unassociated detections that are in the same spatial vicinity. This local matching uses the Hungarian algorithm with a cost matrix based on the Euclidean distance:
    \begin{equation}
    C_{\text{Euc}}(i, j) = \| d^{i,t}_p - \tau^{j,t-1}_p \|.
    \end{equation}
    \item \textbf{Iteration:} New matches are added to \( \mathcal{T}^t_{\text{matched}} \), and the process is repeated, propagating associations outwards from the initial high-confidence matches. This continues until no more associations can be made.
\end{enumerate}

This cascade approach prevents incorrect associations that might arise from noisy GPS data or other outliers, as it relies on the established local structure of the orchard map.

In the case where the initial IoU-based association yields no matches (\( \mathcal{T}^t_{\text{matched}} \) is empty), we perform a global association on all unassociated tracks and detections using the Euclidean distance cost matrix \( C_{\text{Euc}} \) (Stage 2b). This is often the case after a U-turn at the end of a row.

Finally, after all association stages are complete, any detections that remain unmatched are used to initialize new tracks, allowing the map to be incrementally updated as the robot explores the orchard.

\subsubsection{Factor graph}
Tree-SLAM uses a factor graph for the SLAM problem. A factor graph is a type of probabilistic graphical model that represents variables, such as robot poses and landmark positions, and the relationships between them, called factors. In our case, the factor graph is built and optimized using GTSAM\nomenclature{GTSAM}{Georgia Tech Smoothing and Mapping} with the iSAM2\nomenclature{iSAM2}{incremental Smoothing and Mapping} backend, which allows for efficient, incremental updates as new data arrives \cite{dellaert_borglabgtsam_2022}.

\begin{figure}[ht]
    \centering
    \includegraphics[width=0.7\textwidth]{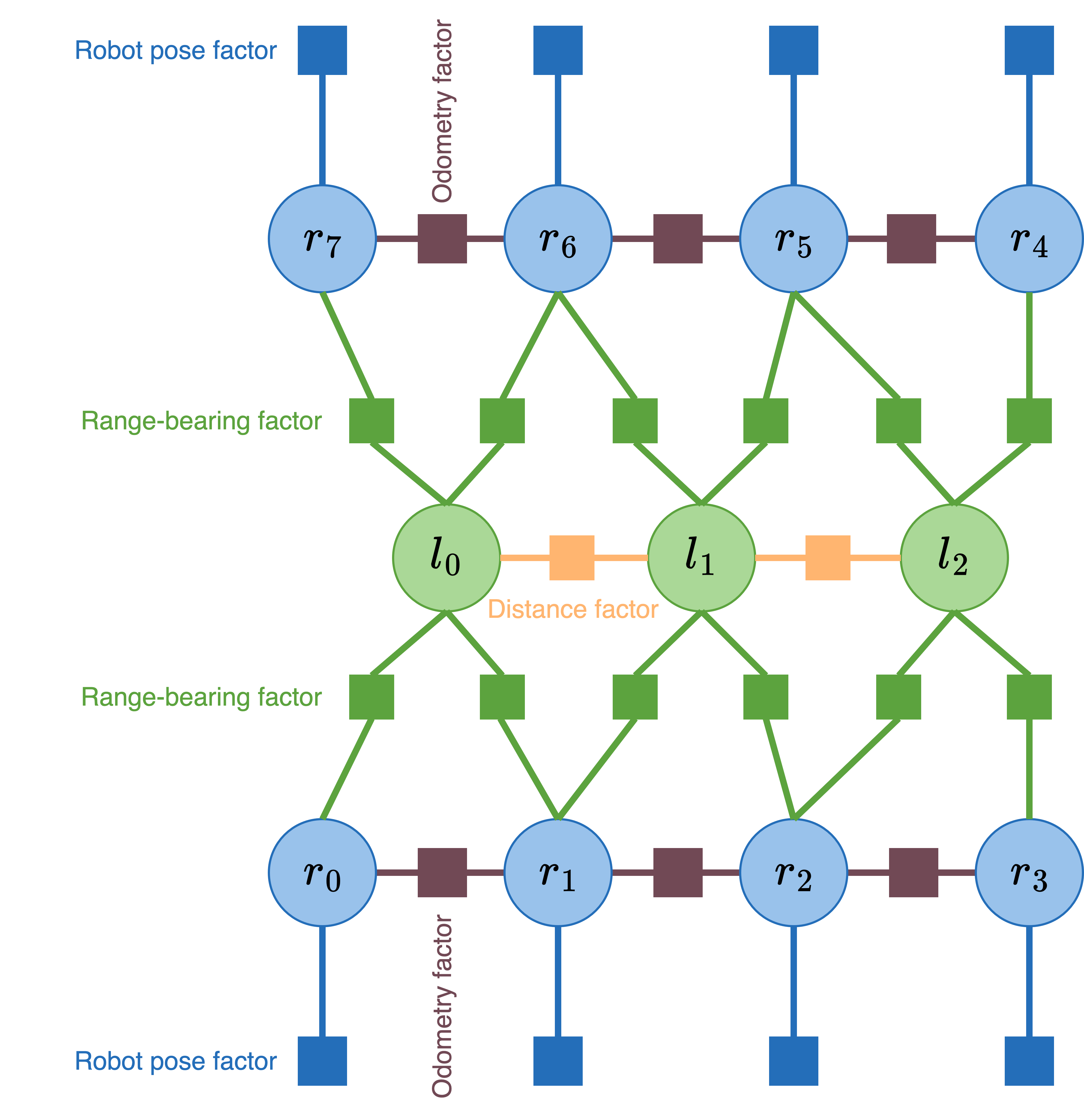}
    \caption{An illustration of the factor graph used in Tree-SLAM. The graph represents the robot's trajectory and the map of the orchard. The variables (circles) are the robot poses at different time steps (\(x_0, \dots, x_7\)) and the 2D positions of the tree landmarks (\(l_0, l_1, l_2\)). The factors (squares) represent constraints from sensor measurements: odometry factors connect consecutive robot poses, GPS factors provide absolute pose measurements, landmark factors connect poses to the trees they observe, and distance factors constrain the relative positions of trees seen simultaneously.}
    \label{fig:fg}
\end{figure}

Our algorithm works in a 2D environment, which is a top-down view of the orchard. This is a reasonable assumption because the ground is mostly flat and the trees are planted in rows. The structure of the factor graph is shown in Figure \ref{fig:fg}.

At each time step \( t \), we add a robot pose variable \( \mathbf{x}_t \) to the graph. This variable represents the robot's 2D position and orientation $\mathbf{x}_t = (x_t, y_t, \theta_t)$, where \( x_t \) and \( y_t \) are the robot's coordinates on the ground, and \( \theta_t \) is its heading.

We connect these pose variables using two types of factors:
\begin{enumerate}
    \item \textbf{Odometry factor}: This factor links consecutive robot poses using the robot's wheel odometry (measured by the Husky's wheel encoders). The odometry provides a relative motion measurement between \( \mathbf{x}_{t-1} \) and \( \mathbf{x}_t \):
    \begin{equation}
    \mathbf{z}_{\text{odom},t} = \mathbf{x}_{t} \ominus \mathbf{x}_{t-1} + \boldsymbol{\epsilon}_{\text{odom}}
    \end{equation}
    where \( \ominus \) is the pose difference operator and \( \boldsymbol{\epsilon}_{\text{odom}} \) is Gaussian noise with covariance \( \mathbf{\Sigma}_{\text{odom}} \).
    \item \textbf{GPS pose factor}: When available, GPS provides an absolute measurement of the robot's pose:
    \begin{equation}
    \mathbf{z}_{\text{GPS},t} = \mathbf{x}_t + \boldsymbol{\epsilon}_{\text{GPS}}
    \end{equation}
    where \( \boldsymbol{\epsilon}_{\text{GPS}} \) is the GPS measurement noise with covariance \( \mathbf{\Sigma}_{\text{GPS}} \).
\end{enumerate}

For mapping, we add landmark variables \( \mathbf{l}_j = (x_{l_j}, y_{l_j}) \) to the graph, where each landmark variable corresponds to a track from the data association step. Each landmark represents a tree's position in 2D world coordinates, and the factor graph stores and maintains only the position of each track.

When the robot detects a tree trunk, the detection can be associated with a landmark ID. The robot observes each landmark from its current pose using range and bearing measurements. The measurement model is:
\begin{equation}
\begin{aligned}
r_{t,j} &= \sqrt{ (x_{l_j} - x_t)^2 + (y_{l_j} - y_t)^2 } + \epsilon_{r}, \\
\phi_{t,j} &= \tan^{-1}(y_{l_j} - y_t, x_{l_j} - x_t) - \theta_t + \epsilon_{\phi},
\end{aligned}
\end{equation}
where \( r_{t,j} \) is the distance from the robot to the tree, \( \phi_{t,j} \) is the angle to the tree relative to the robot's heading, and \( \epsilon_{r} \), \( \epsilon_{\phi} \) are measurement noises.

These measurements are added to the factor graph as \textbf{range-bearing factors}, connecting the robot pose and the landmark.

Additionally, we introduce a \textbf{distance factor} between trees detected in the same frame. Since the relative distance between tree trunks observed by the robot can be measured with high accuracy from the RGB-D data in each frame, we add a factor between each pair of landmarks detected in the same frame. This factor constrains the estimated distance between those tree landmarks to match the measured distance from the sensor:
\begin{equation}
\delta_{t,ij} = \| \mathbf{d}^{i,t}_p - \mathbf{d}^{j,t}_p \| + \epsilon_{\delta}
\end{equation}
where \( \delta_{t,ij} \) is the measured distance between detection positions \( \mathbf{d}^{i,t}_p \) and \( \mathbf{d}^{j,t}_p \) of trunks associated with landmarks \( i \) and \( j \) at time \( t \), and \( \epsilon_{\delta} \) is the measurement noise. This factor then constrains the estimated landmark positions \( \mathbf{l}_i \) and \( \mathbf{l}_j \) to be consistent with this measured distance.

The complete factor graph combines all robot poses, landmarks, and the factors. The goal is to find the best estimates for all variables by minimizing the following cost function:
\begin{equation}
\begin{aligned}
\min_{\{\mathbf{x}_t\}, \{\mathbf{l}_j\}} \quad & \sum_{t} \left\| \mathbf{z}_{\text{odom},t} - (\mathbf{x}_{t} \ominus \mathbf{x}_{t-1}) \right\|_{\mathbf{\Sigma}_{\text{odom}}^{-1}}^2 \\
& + \sum_{t} \left\| \mathbf{z}_{\text{GPS},t} - \mathbf{x}_t \right\|_{\mathbf{\Sigma}_{\text{GPS}}^{-1}}^2 \\
& + \sum_{t,j} \left( \left\| r_{t,j} - \hat{r}_{t,j}(\mathbf{x}_t, \mathbf{l}_j) \right\|_{\sigma_r^{-2}}^2 + \left\| \phi_{t,j} - \hat{\phi}_{t,j}(\mathbf{x}_t, \mathbf{l}_j) \right\|_{\sigma_\phi^{-2}}^2 \right) \\
& + \sum_{t, i < j} \left\| \delta_{t,ij} - \| \mathbf{l}_i - \mathbf{l}_j \| \right\|_{\sigma_\delta^{-2}}^2,
\end{aligned}
\end{equation} 
where \( \hat{r}_{t,j} \) and \( \hat{\phi}_{t,j} \) are the predicted range and bearing based on the current estimates, and the last term incorporates the inter-tree distance factors.
This cost function represents the sum of squared errors between the observed sensor measurements and the predicted values given the current estimates of robot poses and landmark positions. Each term in the sum is weighted by the inverse of its corresponding covariance matrix, such as $\mathbf{\Sigma}_{\text{odom}}^{-1}$ for odometry or $\sigma_r^{-2}$ for range. This weighting reflects the confidence in each measurement: measurements with lower uncertainty have a higher influence on the optimization, while noisier measurements contribute less. By minimizing this weighted sum, the factor graph framework finds the most likely configuration of robot poses and landmark positions that best explains all sensor data, taking into account the reliability of each source of information.

By using iSAM2 in GTSAM, the factor graph is updated and optimized in real-time as new measurements arrive. This results in accurate estimates of both the robot's path and the positions of all mapped trees.

\section{Experimental evaluation and results}
\subsection{Tree detection performance}
Accurate and robust tree detection is an important component of the proposed Tree-SLAM pipeline. We evaluated the YOLOv8 instance segmentation model on the labeled test images from the datasets described in Table \ref{tab:dataset}. To analyze the impact of tree type and seasonal variation, we grouped the results by species (pear and apple) and by season (leafless and leafed), which represent the main sources of variability in our data.

Table \ref{tab:detection_perf} summarizes the detection performance in terms of box and mask mAP50. For pear trees, the model achieved consistently high performance across both seasons, with box mAP50 values above 0.9. This robustness can be attributed to the larger size and more defined structure of older pear trunks, which provide distinctive visual features regardless of foliage. In contrast, detection performance for apple trees was more sensitive to seasonal changes. In leafless conditions, the model achieved high accuracy (box mAP50 of 0.98), but performance dropped in leafed conditions (box mAP50 of 0.79). This decline is primarily due to increased occlusion from grass and weeds, which is more pronounced for the younger and thinner apple trees.

Overall, the results highlight the robustness of YOLOv8 in detecting older and thicker trees under varying conditions but underscore challenges in detecting young trees, particularly in cluttered leafed scenes. 

\begin{table}[htbp]
\centering
\caption{YOLOv8 instance segmentation performance on the test set. The table shows mean Average Precision at an IoU threshold of 0.5 (mAP50) for both bounding box and segmentation mask predictions, broken down by tree type and season.}
\begin{tabular}{cccc}
\hline
Dataset         & Season    & Box mAP50 & Mask mAP50 \\ \hline
\multirow{2}{*}{Pear} & Leafless    & 0.97     & \textbf{0.98}      \\
                            & Leafed    & 0.90     & 0.92      \\
\multirow{2}{*}{Apple}   & Leafless    & \textbf{0.98}     & \textbf{0.98}      \\
  & Leafed    & 0.79     & 0.81      \\ \hline
\end{tabular}
\label{tab:detection_perf}
\end{table}

\subsection{Mapping accuracy}
To evaluate the geo-localization accuracy of individual trees, we used ground truth GPS positions collected for multiple trees across several rows, as detailed in Table \ref{tab:dataset}. We assessed the performance of our mapping algorithm using several metrics: the mean Euclidean distance for true positive matches, precision, recall, F1-score, and the percentage of ground truth trees found within half the planting distance. To compute these metrics, we first established a correspondence between the set of predicted trees and the set of ground truth trees using a minimum-cost matching algorithm based on Euclidean distance. An association was considered a true positive (TP)\nomenclature{TP}{True Positive} only if the distance was below a gating threshold. For our main evaluation, we set this threshold to half the planting distance ($PD/2$)\nomenclature{$PD$}{Planting Distance} to filter out incorrect associations. The mean Euclidean distance was then calculated for all TPs. Precision, recall and F1-score were derived from the counts of TPs, false positives (FP\nomenclature{FP}{False Positive}, unmatched predictions), and false negatives (FN\nomenclature{FN}{False Negative}, unmatched ground truth trees). The percentage of trees within half the planting distance corresponds to the recall at the $PD/2$ threshold.

\begin{figure}[ht]
    \centering
    \includegraphics[width=1\textwidth]{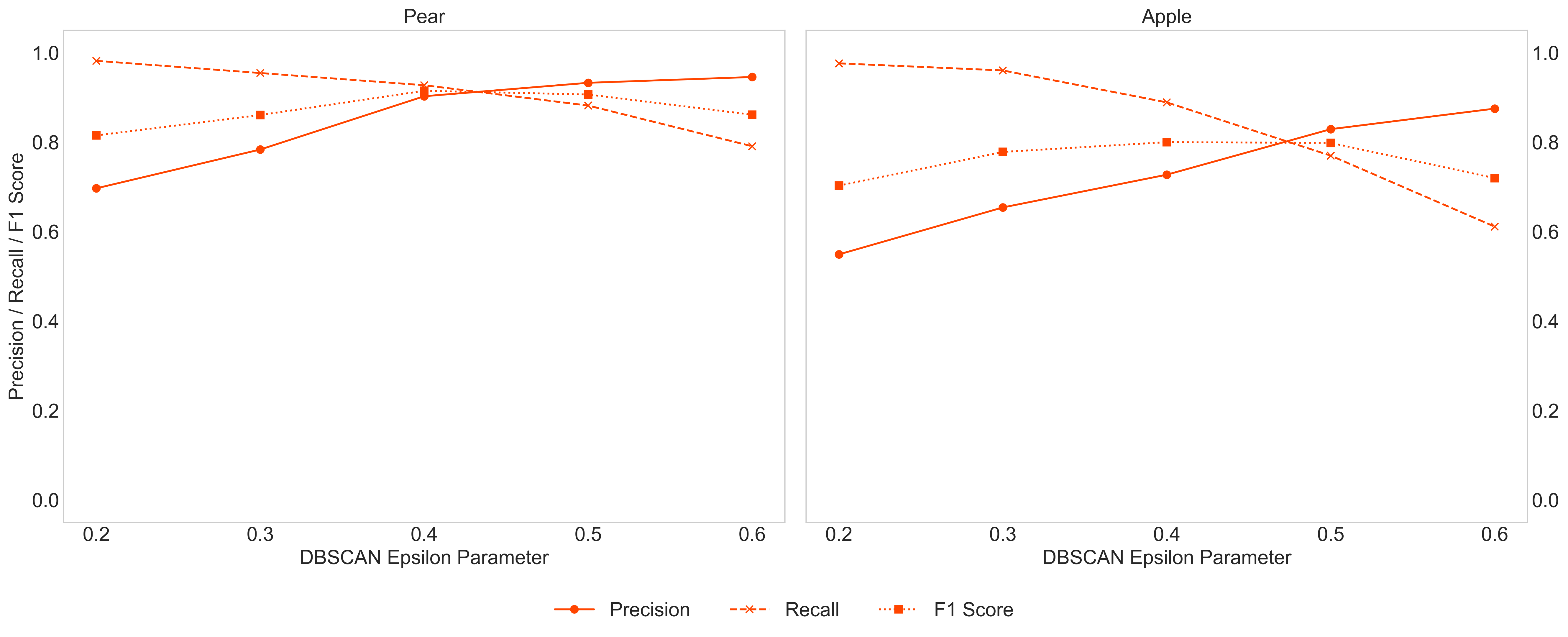}
    \caption{Selection of the optimal epsilon parameter for the DBSCAN-based baseline method. The figure plots precision, recall, and F1-score against different epsilon values for both pear (left) and apple (right) orchards. The F1-score peaks in the 0.4-0.5 meter range, indicating the optimal setting for clustering tree detections.}
    \label{fig:pr_curves_dbscan}
\end{figure}

We compare our Tree-SLAM algorithm against a baseline method based on the work of \citep{brown_tree_2024}, which relies on clustering detections. This baseline uses the same YOLOv8 trunk detection model but does not use the factor graph for data association and pose refinement. Instead, it accumulates all trunk detections throughout a run and applies the DBSCAN clustering algorithm to the entire set of 2D positions. The centroid of each resulting cluster is then considered the final estimated position of a tree. To determine the optimal DBSCAN parameters, we evaluated its performance across a range of epsilon values, as shown in Figure \ref{fig:pr_curves_dbscan}. The results indicate that an epsilon of 0.4 to 0.5 meters, which is slightly less than half the planting distance, yields the best F1-score. Based on these results, an epsilon of 0.5 was used for the baseline in the following experiments. The minimum number of samples per cluster was set to 5, as varying this parameter showed negligible impact on performance.

\begin{figure}[ht]
    \centering
    \includegraphics[width=1\textwidth]{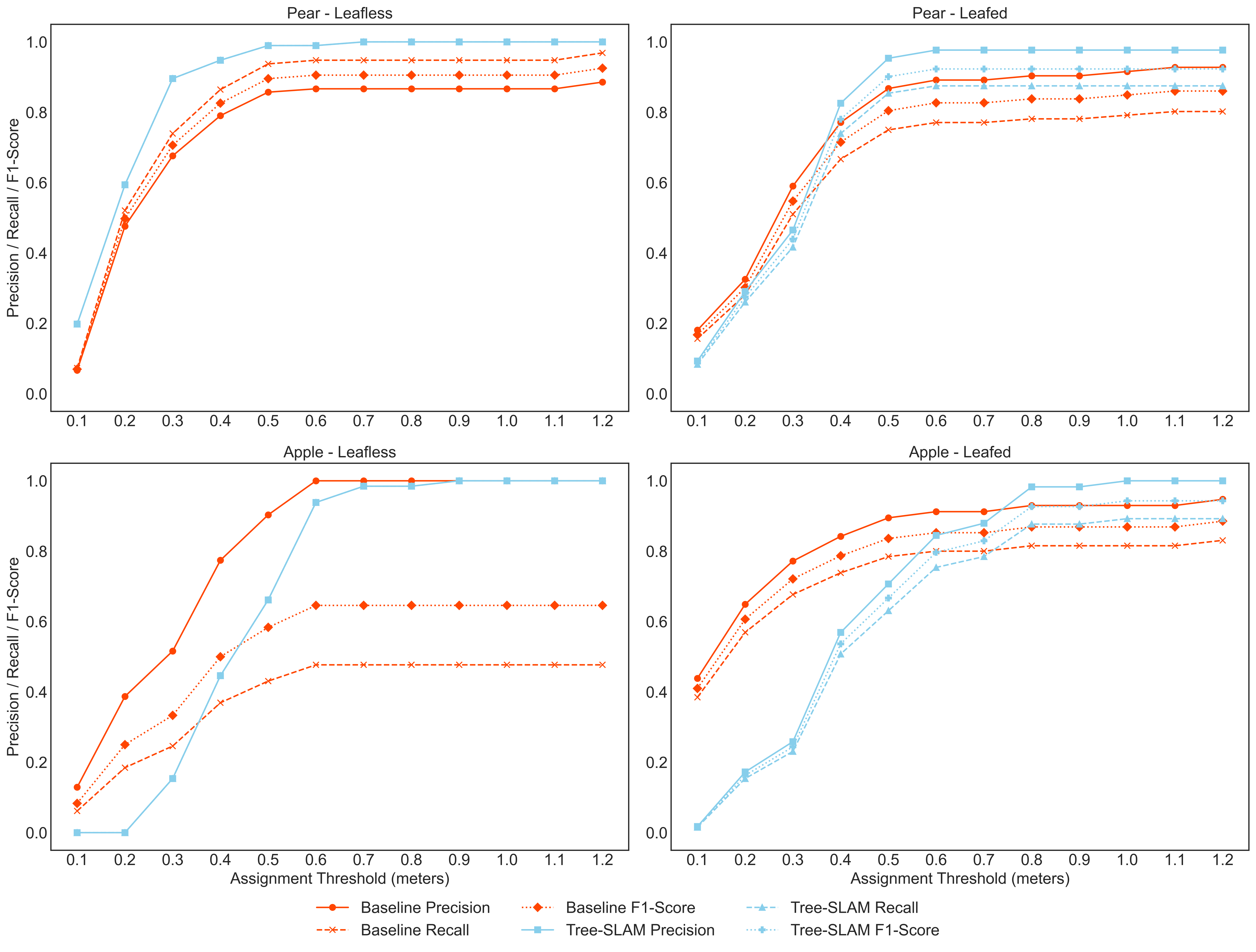}
    \caption{Comparison of mapping performance between Tree-SLAM and the baseline method across different datasets and seasons. The plots show precision, recall, and F1-score as a function of the assignment threshold (Euclidean distance). The four subplots show results for pear and apple orchards in both leafless and leafed conditions.}
    \label{fig:pr_curves}
\end{figure}

\begin{figure}[ht]
    \centering
    \includegraphics[width=1\textwidth]{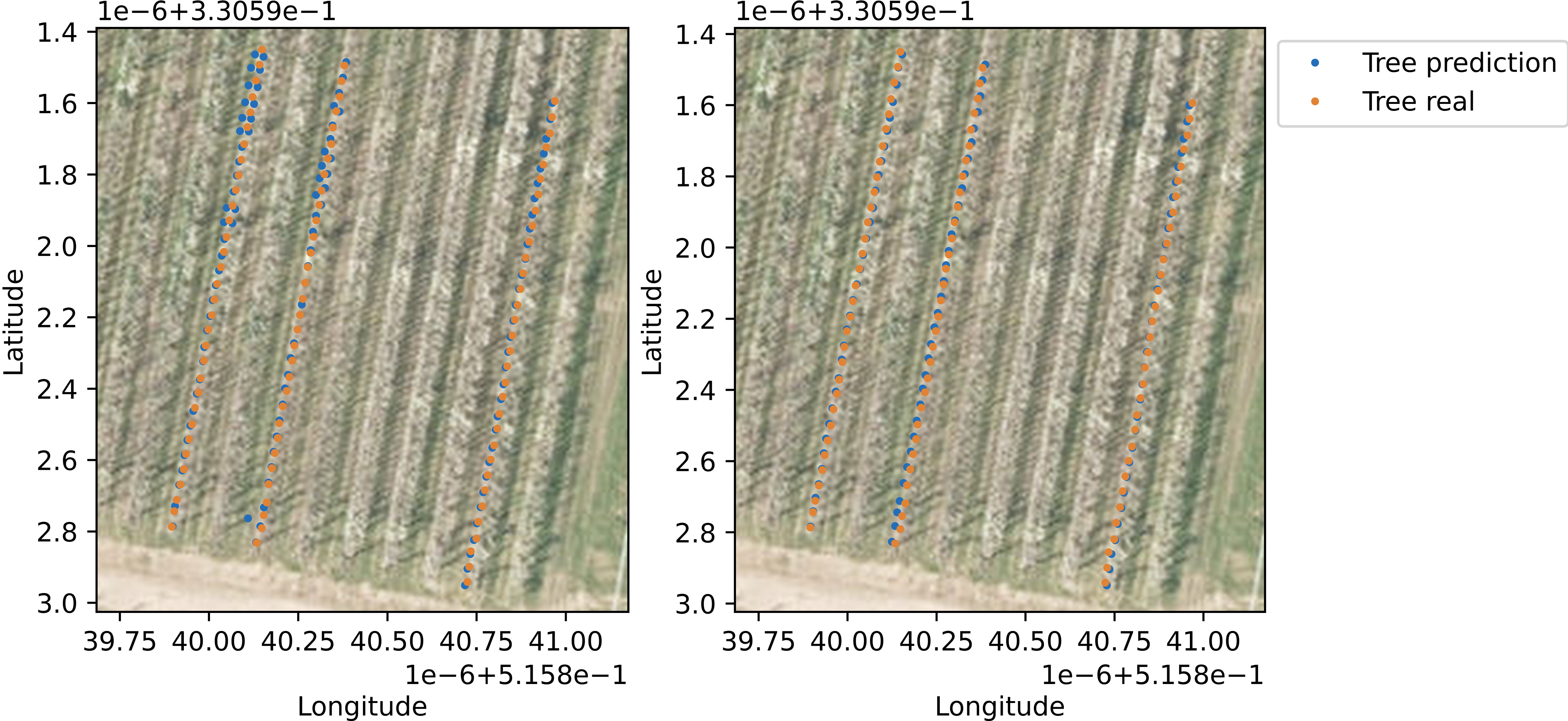}
    \caption{Satellite view containing ground truth and predicted tree positions for the pear trees during the leafless recordings. The left plot shows the predictions of the baseline algorithm while the right one displays the predictions of our Tree-SLAM algorithm.}
    \label{fig:sat}
\end{figure}

Figure \ref{fig:pr_curves} shows the precision, recall and F1-score curves for both methods across a range of Euclidean gating thresholds. The curves demonstrate that Tree-SLAM consistently achieves higher precision, recall and F1-score than the baseline across most scenarios and around the $PD/2$ threshold (0.55 m for pear trees and 0.6 m for apple trees). This confirms that our method is not only finding more trees but is also more precise in its predictions. The only exception is for apple trees in leafed conditions, where the baseline shows slightly better performance around the $PD/2$ threshold. This can be attributed to the lower trunk detection performance in leafed conditions for young apple trees as seen in Table \ref{tab:detection_perf}, which negatively impacts the data association and landmark updates in our more complex SLAM pipeline, while the simpler clustering of the baseline is less affected by intermittent detection failures. Furthermore, it can be seen that both the leafless results for apple and pear trees show the same precision and recall values for the Tree-SLAM algorithm. This is because our algorithm does not predict trees that do not actually exist: all trees can be matched to a ground truth tree when increasing the distance threshold. This is not the case for the baseline algorithm, which predicts more trees than are actually present in the orchard, leading to lower precision and recall. This can be seen in the satellite view image of Figure \ref{fig:sat}, where the baseline algorithm predicts more trees than the ground truth, while Tree-SLAM only predicts the trees that were actually measured.

\begin{table}[htbp]
\centering
\caption{Comparison of mapping accuracy between the proposed Tree-SLAM algorithm ('Ours') and the DBSCAN-based baseline. The table shows the percentage of trees correctly identified within half the planting distance ($PD/2$) and the mean localization error for true positive (TP) detections, categorized by tree type and season.}
\resizebox{\linewidth}{!}{
\begin{tabular}{ccccc}
\hline
\textbf{Row} & \textbf{Algorithm} & \textbf{Season} & \textbf{\% within $PD/2$}$\uparrow$ & \textbf{$TP$ mean Error (m)}$\downarrow$ \\ \hline
\multirow{4}{*}{Pear} & Baseline & Leafless & 94.8\% & 0.22 \\
& Ours & Leafless & \textbf{99.0\%} & 0.18 \\
& Baseline & Leafed & 77.1\% & 0.24 \\
& Ours & Leafed & 85.4\% & 0.28 \\ \hline
\multirow{4}{*}{Apple} & Baseline & Leafless & 44.6\% & 0.26 \\
& Ours & Leafless & 84.6\%& 0.40 \\
& Baseline & Leafed & 80.0\% & \textbf{0.16} \\
& Ours & Leafed & 69.2\% & 0.32 \\ \hline
\end{tabular}
}
\label{tab:no_fg}
\end{table}

Table \ref{tab:no_fg} presents a detailed comparison of Tree-SLAM and the baseline algorithm across different tree types and seasons, using the percentage of correctly identified trees within half the planting distance and the mean error for these true positives. Our proposed method generally outperforms the baseline in terms of \% within PD/2 (recall), with the exception of apple trees in leafed conditions, which is probably due to the lower detection performance as stated earlier. With a lower overall detection rate, the number of noisy measurements is also reduced, which in turn decreases the likelihood of incorrect clustering in the baseline method. For pear trees, Tree-SLAM improves recall from 94.8\% to 99.0\% in leafless conditions and from 77.1\% to 85.4\% in leafed conditions. The advantage is particularly significant for apple trees in leafless conditions, where Tree-SLAM achieves 84.6\% recall against the baseline's 44.6\%. This demonstrates the robustness of our factor graph approach in GPS-constrained scenarios. The baseline, which relies solely on clustering raw detections, struggles when GPS inaccuracies introduce significant noise, leading to a failure in differentiating tree clusters and resulting in many missed trees as shown in the satellite view \ref{fig:sat}.

In terms of mean error for correctly identified trees, the results are more mixed. Tree-SLAM achieves a lower error for pear trees in leafless conditions: 0.18 m against 0.22 m. However, the baseline achieves a lower mean error in the other three scenarios, for example, with apple trees in leafed conditions. This is likely due to the baseline predicting more trees than actually exist by creating multiple clusters for the same tree, which tends to make one of the clusters more accurate. In other words, this lower error tends to come at the cost of lower precision and recall for the baseline.



\subsection{Ablation studies}
To understand how different parts of our algorithm contribute to the final performance, we conducted ablation studies. These studies involved removing or changing relevant components to see their effect. We focused on two main parts: the PCA-based trunk center estimation and the graph-based data association. We tested two variations of our algorithm:

\begin{enumerate}
    \item \textbf{Without PCA-based Trunk Center Estimation}: We replaced the PCA-based trunk center estimation with a simpler averaging method. This simpler method calculates the trunk center as the average of all its detected points, without considering their spatial layout as PCA does.
    \item \textbf{Without Graph-based Data Association}: We removed the graph-based data association stage (stage 2a in \ref{fig:association}). In this version, the algorithm tried to associate all detected trunks without using the neighborhood information and spatial rules that the graph method provides.
\end{enumerate}

We compared the mapping accuracy of these two variations against our complete algorithm. The tests were run under the same conditions for a fair comparison. Table \ref{tab:no_pca} shows the results of these studies.

The results show that PCA-based trunk center estimation is very important for mapping the older pear trees. As seen in Table \ref{tab:no_pca}, removing PCA for pear trees significantly lowered precision from 0.99 to 0.79 in leafless conditions and from 0.98 to 0.86 in leafed conditions. The simpler averaging method is less accurate for these larger trunks, leading to more association errors. The opposite was true for the young apple trees in leafed conditions. For this case, removing PCA improved both precision and recall. This is likely because the small apple trunks were surrounded by protective nets and tall grass, as shown in Figure \ref{fig:tree_examples}. This created noisy point clouds where the simple averaging method performed better than the more complex PCA estimation.

Our graph-based data association method also proved to be a critical component. Its importance was most clear in the challenging leafed conditions for apple trees. Removing the graph-based stage caused the recall to drop dramatically from 0.75 to 0.42 and the mean error to increase from 0.34 m to 0.47 m. This demonstrates that using the spatial relationships between trees is essential for correct data association when individual detections are less reliable. For the easier cases, like pear trees in leafless conditions, the graph-based method mainly helped to improve the final tree positions, reducing the mean error from 0.30 m to 0.18 m without changing the high precision and recall.

Overall, these studies confirm that both PCA-based center estimation and graph-based data association are important to the algorithm's performance. They significantly improve the mapping accuracy and robustness, especially in difficult scenarios with visual noise or challenging foliage.

\begin{table}[htbp]
\centering
\caption{Ablation study on the impact of PCA based trunk center estimation and graph based data association. The table compares the full algorithm against versions with each component disabled, evaluated on pear and apple datasets across seasons.}
\resizebox{\linewidth}{!}{
    \begin{tabular}{cccccccc} 
    \hline
    \textbf{Row} & \textbf{PCA} & \textbf{Graph Association} & \textbf{Season} & \textbf{$TP$ mean Error (m)}$\downarrow$ & \textbf{Precision}$\uparrow$ & \textbf{Recall}$\uparrow$ \\ \hline
    \multirow{6}{*}{Pear} & \checkmark & \checkmark & Leafless & \textbf{0.18} & \textbf{0.99} & \textbf{0.99} \\
    & & \checkmark & Leafless & 0.33 & 0.79 & 0.97 \\
    & \checkmark & & Leafless & 0.30 & \textbf{0.99} & \textbf{0.99} \\
    & \checkmark & \checkmark & Leafed & 0.29 & 0.98 & 0.88 \\
    & & \checkmark & Leafed & 0.32 & 0.86 & 0.80 \\
    & \checkmark & & Leafed & 0.35 & 0.95 & 0.85 \\ \hline
    \multirow{6}{*}{Apple} & \checkmark & \checkmark & Leafless & 0.42 & 0.94 & 0.94 \\
    & & \checkmark & Leafless & 0.40 & 0.82 & 0.82 \\
    & \checkmark & & Leafless & 0.37 & 0.94 & 0.94 \\
    & \checkmark & \checkmark & Leafed & 0.34 & 0.85 & 0.75 \\
    & & \checkmark & Leafed & 0.35 & 0.95 & 0.89 \\
    & \checkmark & & Leafed & 0.47 & 0.90 & 0.42 \\ \hline
    \end{tabular}
}
\label{tab:no_pca}
\end{table}

\section{Discussion}
This study introduced Tree-SLAM, a semantic SLAM system designed to generate accurate geo-localized maps of individual trees in orchards. Our approach demonstrates robust performance across different seasons and tree types, achieving a mapping accuracy as low as 18 cm, which is less than 20\% of the typical planting distance. This level of precision is critical for enabling autonomous, tree-level agricultural operations.

The first step of our system is a trunk detection model. Our YOLOv8-based detector achieved a box mAP50 of up to 0.98, outperforming the 0.82 reported by \citep{huang_deep-learning-based_2023} using YOLOv5, although their front-facing camera setup presents a different challenge. Our results are more comparable to the 0.99 mAP50 achieved by \citep{brown_tree_2024} with a similar side-facing camera configuration. A strength of our model is its seasonal robustness, particularly for mature pear trees. However, its performance declined for young apple trees in leafed conditions, primarily due to heavy occlusion from grass and weeds, a common challenge for vision-based systems in agricultural settings.

One of the main contributions of this work lies in the SLAM framework. When compared to a baseline method that clusters raw detections \cite{brown_tree_2024}, Tree-SLAM demonstrated significantly higher recall and F1-scores across most scenarios. The baseline's performance degraded substantially in the presence of noisy GPS data, particularly for the apple orchard, where it failed to correctly cluster and identify a large portion of the trees. In contrast, our factor graph approach effectively integrates noisy GPS, odometry, and landmark observations, correcting for sensor inaccuracies and resulting in a more complete and precise map, as visualized in Figure \ref{fig:sat}. This highlights the necessity of a tightly-coupled sensor fusion approach for reliable mapping in real-world orchard conditions where GPS signals are often unreliable.

The ablation studies confirmed the importance of our proposed components. The PCA-based trunk center estimation was shown to be important for accurately localizing mature pear trees but was less effective for young, occluded apple trees where simpler averaging of the noisy point cloud yielded better results. This suggests that the optimal method for estimating the trunk center depends highly on the specific characteristics of the trees and their environment. Furthermore, the cascade-graph data association was important for maintaining track identities in challenging conditions, such as the cluttered leafed apple orchard with noisy GPS data. By leveraging the known spatial structure of the orchard, it prevented incorrect associations and significantly improved recall.

While our system yields accurate mapping results, the performance drop in cluttered leafed conditions for young trees represents a limitation and an area for future improvement. Integrating other sensor modalities, such as LiDAR, could provide more robust geometric information, mitigating issues of visual occlusion. Nonetheless, Tree-SLAM provides a robust and accurate solution for orchard mapping, taking a step toward advanced precision agriculture applications like automated individual tree monitoring and targeted interventions.

\section{Conclusion and future work}
This work presents a semantic SLAM algorithm designed for agricultural robots to accurately map and geo-localize trees in orchards, even under challenging GPS conditions. By integrating multi-sensor data through a factor graph framework and leveraging novel trunk detection and data association methods, Tree-SLAM achieves a mapping error lower than 20\% of the tree planting distance. The importance of the individual components, such as the PCA-based trunk localization and graph-based data association, was confirmed through ablation studies, which showed their significant contribution to the overall performance. Results showed better precision and recall values in the resulting tree map compared to a recent baseline method, especially in conditions with low GPS accuracy. 

Future work will focus on enhancing the system's capabilities in two main areas. First, we will explore the integration of LiDAR data to improve robot localization performance and more accurate measurements of tree locations. Second, we will deploy the algorithm on a fully autonomous UGV to demonstrate how precise mapping enables targeted agricultural tasks. Third, we will research active perception strategies \cite{burusa_semantics-aware_2024}, allowing the robot to autonomously re-observe trees with high uncertainty, thereby improving overall map accuracy.

\section*{CRediT author statement}
\textbf{David Rapado-Rincon}: Conceptualization, Methodology, Software, Investigation, Data Curation, Writing - Original draft; \textbf{Gert Kootstra}: Conceptualization, Writing - Review \& Editing, Supervision, Funding acquisition.

\section*{Funding}
This research is part of the UTOPIA project. It has received funding from the European Union’s Horizon 2020 research and innovation programme under grant agreement Number 862665 ERA-NET ICT-AGRI-FOOD. UTOPIA is part of the ERA-NET Cofund ICT-AGRI-FOOD, with funding provided by national sources FWO, TUBITAK, Dutch ministry of Agriculture, Nature and Food Quality and co-funding by the European Union’s Horizon 2020 research and innovation program, Grant Agreement number 862665.

\section*{Declaration of competing interest}
We declare that there are no personal and/or financial relationships that have inappropriately affected or influenced the work presented in this paper.

\bibliographystyle{elsarticle-num}  
\bibliography{references}

\end{document}